\documentclass[runningheads]{llncs}

\usepackage{eccv}

\usepackage{eccvabbrv}

\usepackage{graphicx}
\usepackage{booktabs}
\usepackage{wrapfig}

\usepackage[accsupp]{axessibility}  

\usepackage{hyperref}

\usepackage{orcidlink}

\usepackage{algorithm}
\usepackage{algorithmicx}
\usepackage{algpseudocode}

\usepackage{graphicx}
\usepackage{float}
\usepackage{subcaption}
\usepackage{multirow}   
\usepackage{caption}
\usepackage{xcolor}
\usepackage{pifont} 
\usepackage[table]{xcolor}

\newcommand{\equalcontrib}{\textsuperscript{*}}   
\newcommand{\corrauthor}{\textsuperscript{\textdagger}} 

\begin{document}

\title{RT-RMOT: A Dataset and Framework for \\ RGB-Thermal Referring Multi-Object Tracking} 

\titlerunning{A Dataset and Framework for RGBT RMOT}

\author{
Yanqiu Yu\inst{1}\equalcontrib \and
Zhifan Jin\inst{2}\equalcontrib \and
Sijia Chen\inst{1}\equalcontrib \and
Tongfei Chu\inst{1} \and
En Yu\inst{1} \and
Liman Liu\inst{2} \and
Wenbing Tao\inst{1}\corrauthor
}

\authorrunning{Yu et al.} 

\institute{
Huazhong University of Science and Technology, Wuhan, Hubei, China \\
\email{\{yanqiuyu6, sijiachen, tongfeichu, yuen, wenbingtao\}@hust.edu.cn} 
\and
South-Central Minzu University, Wuhan, Hubei, China \\
\email{\{zhifanjin, limanliu\}@mail.scuec.edu.cn}
}

\maketitle

\begingroup
\renewcommand\thefootnote{}
\footnotetext{* Equal contribution}
\footnotetext{\textdagger\ Corresponding author}
\endgroup
\begin{abstract}
  \quad Referring Multi-Object Tracking has attracted increasing attention due to its human-friendly interactive characteristics, yet it exhibits limitations in low-visibility conditions, such as nighttime, smoke, and other challenging scenarios. To overcome this limitation, we propose a new \textbf{R}GB–\textbf{T}hermal \textbf{RMOT} task, named \textbf{RT-RMOT}, which aims to fuse RGB appearance features with the illumination robustness of the thermal modality to enable all-day referring multi-object tracking. To promote research on RT-RMOT, we construct the first \textbf{Ref}erring Multi-Object Tracking dataset under \textbf{R}GB-\textbf{T}hermal modality, named \textbf{RefRT}. It contains 388 language descriptions, 1,250 tracked targets, and 166,147 Language-RGB–Thermal (L-RGB–T) triplets. Furthermore, we propose \textbf{RTrack}, a framework built upon a multimodal large language model (MLLM) that integrates RGB, thermal, and textual features. Since the initial framework still leaves room for improvement, we introduce a Group Sequence Policy Optimization (GSPO) strategy to further exploit the model’s potential. To alleviate training instability during RL fine-tuning, we introduce a Clipped Advantage Scaling (CAS) strategy to suppress gradient explosion. In addition, we design Structured Output Reward and Comprehensive Detection Reward to balance exploration and exploitation, thereby improving the completeness and accuracy of target perception. Extensive experiments on the RefRT dataset demonstrate the effectiveness of the proposed RTrack framework. 
  
\end{abstract}

\section{Introduction}
\label{sec:Introduction}
\quad With the recent advances in vision–language understanding, Referring Multi-Object Tracking (RMOT) has emerged as a challenging and active research topic due to its human-friendly interactive characteristics. RMOT aims to track specific targets according to language description. Numerous RMOT approaches have appeared, including TransRMOT \cite{wu2023referring}, TempRMOT \cite{zhang2024bootstrapping}, and CPAny \cite{li2025cpany}.

However, current RMOT research is largely constrained to RGB observation modality. In low-visibility conditions, such as nighttime and smoke, the appearance information of targets often degrades severely, making it difficult for the model to establish reliable semantic associations between compromised visual features and language descriptions. RMOT task paradigm inherently limits the robustness of language-guided tracking systems in complex and dynamic environments, making it inadequate for all-day real-world applications. For example, in nighttime, tunnel, or smoky scenarios, RGB images often become dark or noisy and fail to provide clear contour information of the targets, making it difficult for the model to accurately localize targets.

\begin{figure}[t]
    \setlength{\abovecaptionskip}{0.8mm}  
    \setlength{\belowcaptionskip}{-2.4mm}  
    \centering
    \includegraphics[width=1.0\linewidth]{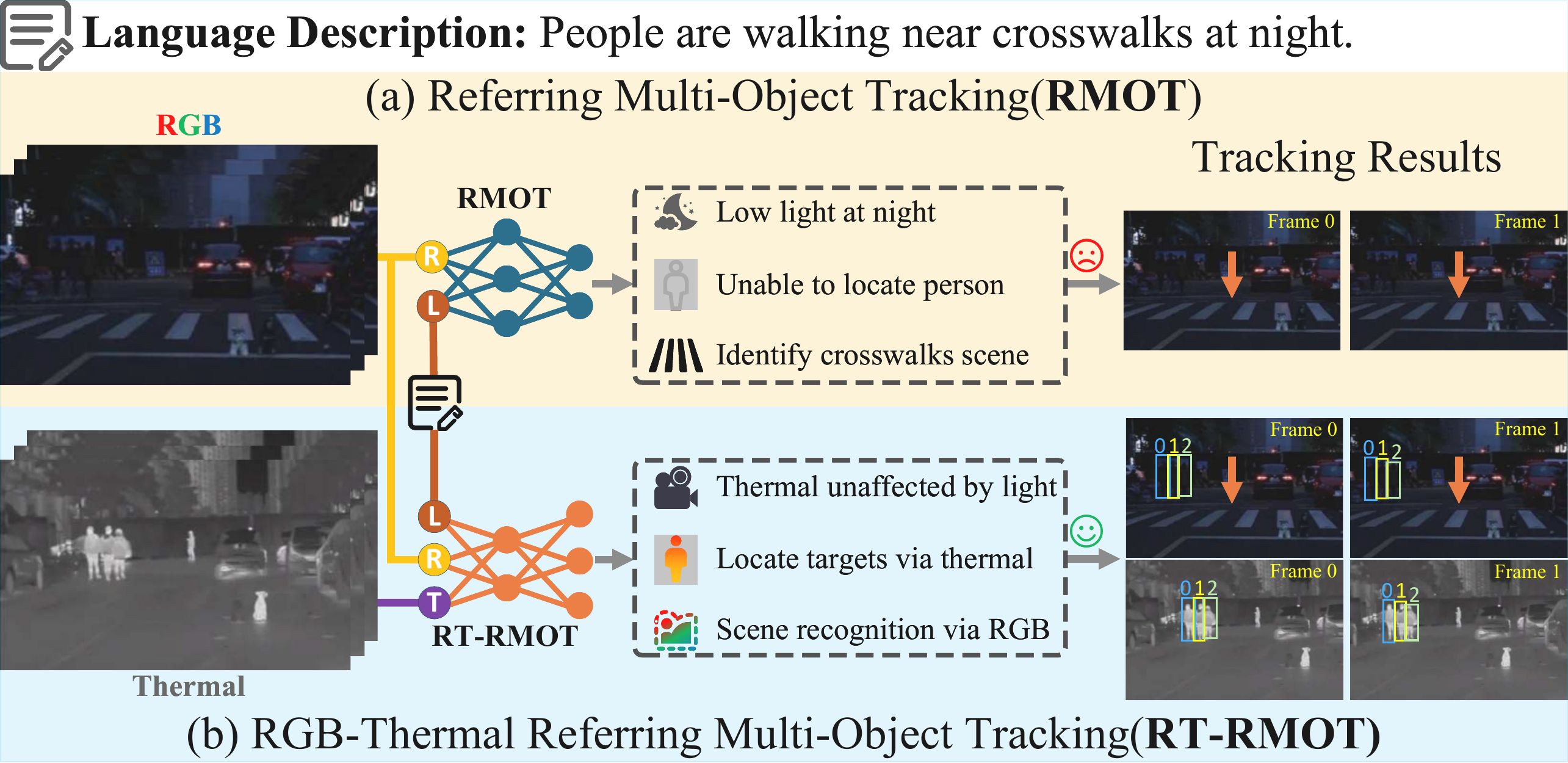}
    \caption{\textbf{The difference between RMOT and RT-RMOT.} RMOT model takes RGB images as input, but pedestrian positions cannot be reliably obtained from RGB alone, often leading to tracking failure. RT-RMOT model leverages thermal (T) images to obtain pedestrian contours and locations, and combines them with crosswalk regions provided by RGB images, enabling precise identification and tracking of people described by language.}
    \label{fig:first figure}
\end{figure}

To overcome this limitation, we propose a new task, \textbf{R}GB-\textbf{T}hermal \textbf{R}eferring \textbf{M}ulti-\textbf{O}bject \textbf{T}racking (\textbf{RT-RMOT}). This task requires the model to deeply fuse language, RGB, and thermal multimodal data, enabling robust all-day referring multi-object tracking even under low-visibility conditions such as nighttime and smoke. In this task, the language modality allows users to specify targets in a human-friendly manner, the RGB modality captures fine-grained appearance details to distinguish between targets, and the infrared modality provides contour information to facilitate accurate localization. Such a integrated multimodal paradigm significantly enhances the reliability of referring multi-object tracking in complex environments. As shown in Fig.~\ref{fig:first figure}, for the language description “people walking near a crosswalk at night”, RT-RMOT task recognizes the crosswalk using RGB images and detects pedestrians using thermal images, enabling accurate identification of the people mentioned in the language description.

To advance the research of RT-RMOT task, we construct the first \textbf{Ref}erring \textbf{R}GB-\textbf{T}hermal Multi-Object Tracking dataset, named \textbf{RefRT}. RefRT dataset consists of 388 high-quality language descriptions, 72 scenes, 1,250 referred targets and 166,147 language-RGB-thermal triplets, covering diverse scenarios such as campuses and urban areas, as well as imaging conditions including night, rain, and snow. It provides a high-quality and adaptable foundation for RMOT research under RGBT scenarios.

To address the challenges of multimodal feature fusion and effective information utilization in the RT-RMOT task, we propose a framework, termed \textbf{RTrack}. Centered on a Multimodal Large Language Model (MLLM) with strong cross-modal alignment and reasoning capabilities, RTrack achieves joint representation learning of language, RGB and thermal features for referring tracking. As the initial framework still leaves room for performance improvement, we further introduce a Group Sequence Policy Optimization (GSPO) \cite{zheng2025group} strategy to better unlock the model’s potential. To ensure stable fine-tuning, we propose a novel Clipped Advantage Scaling (CAS) strategy to effectively suppress gradient fluctuations commonly observed in reinforcement learning. In addition, to balance exploration and exploitation, we carefully design Structured Output and Comprehensive Detection Reward that guides the model to maintain target perception integrity and spatial precision in challenging environments.

Experimental results demonstrate that RTrack significantly achieves state-of-the-art (SOTA) performance. Our method achieves the highest scores across five evaluation metrics: HOTA, DetA, AssA, DetRe, and AssRe. Among these, the HOTA metric provides a comprehensive assessment of both detection and tracking performance. Compared with other methods, our approach improves the HOTA metric by 6.84\%, DetA by 9.8\%, DetRe by 17.1\%, and DetPR by 5.64\%, demonstrating strong generalization capability on the RefRT dataset.  

In summary, the main contributions are as follows:
\begin{itemize}
\item We propose a \textbf{R}GB-\textbf{T}hermal \textbf{R}eferring \textbf{M}ulti-\textbf{O}bject \textbf{T}racking (\textbf{RT-RMOT}) task, which aims to integrate the high-level semantic information from the language modality, the appearance features from the RGB modality, and the contour cues from the thermal modality, thereby overcoming the perceptual limitations of traditional RMOT under low-visibility conditions such as nighttime and smoke, and enabling all-day referring multi-object tracking.

\item We construct \textbf{RefRT}, the first dataset specifically designed for RT-RMOT task. It provides  pixel-level alignment between RGB and thermal modalities, along with high-quality language descriptions. It contains 388 language descriptions, 1,250 referred targets, and 72 scenes.
\item We propose \textbf{RTrack}, a unified framework guided by a Multimodal Large Language Model (MLLM) that integrates language, RGB and thermal information. By effectively exploiting the complementary cues of both the three modalities, RTrack achieves precise cross-modal fusion and all-day tracking. Extensive experiments on the RefRT dataset achieve state-of-the-art (SOTA) performance, highlighting the effectiveness of the framework.
\end{itemize}

\section{Related Work}
\noindent \textbf{Referring Multi-Object Tracking.} It aims to simultaneously track multiple language-specified objects in video. TransRMOT \cite{wu2023referring} introduces the task, proposing an end-to-end Transformer framework that handles cross-modal reference and temporal association. Subsequent works, such as iKUN \cite{du2024ikun} and CPAny \cite{li2025cpany}, focus on optimizing cross-modal fusion and improving tracking paradigms within the dominant Transformer architecture. However, these methods often struggle with dynamic references involving temporal status changes (e.g., ``the moving vehicles''). To enhance robustness against these challenges, TempRMOT \cite{zhang2024bootstrapping} presents a query-driven Temporal Enhancement Module (TEM), which refines object queries using long-term spatiotemporal interaction with historical frames, significantly boosting temporal consistency.

\noindent \textbf{RGB-Thermal Object Tracking.} RGBT tracking integrates RGB and thermal infrared (T) data to improve robustness against challenges such as poor illumination and occlusion. Early methods \cite{long2019multi, zhang2020object, wang2020cross} relied on shallow feature fusion via handcrafted features or correlation filters.
With the advent of deep learning, the focus shifted toward modality-adaptive representation \cite{zhang2023efficient}, utilizing dual-stream networks and dynamic attention or distillation mechanisms to balance modality contributions. Most recently, Transformer architectures \cite{xiao2025cross} have been adopted for their global modeling capacity to address modality heterogeneity and long-range dependencies. Notably, DeformCAT \cite{hu2025deformle} introduces a deformable cross-attention mechanism that flexibly samples cross-modal features, significantly enhancing robustness to target shape and positional variations.

\noindent \textbf{Multimodal Large Language Models.} They have rapidly progressed from simple alignment to complex multimodal reasoning. Foundational work like CLIP \cite{radford2021learning} established contrastive learning for image–text alignment. The shift toward reasoning was pioneered by LLaVA \cite{liu2023visual}, which integrated Multimodal Large Language Models (MLLMs) via instruction tuning. Subsequent powerful models, including GPT-4 \cite{achiam2023gpt} and Gemini \cite{team2023gemini}, demonstrated unified perception and reasoning. Recent efforts, such as Qwen2.5-VL \cite{bai2025qwen2} and InternVL2 \cite{chen2024internvl}, focus on enhancing fine-grained visual grounding, video comprehension, and overall generalization capacity through efficient fusion techniques.

\section{Dataset}
\quad To advance research on the RGB-Thermal Referring Multi-Object Tracking (RT-RMOT) task, address the scarcity of datasets in this field, and evaluate model performance, we construct the RefRT dataset.

\subsection{Dataset Collection}
\label{subsubsec:Dataset Collection}
\quad Our task requires the source dataset to have three core characteristics: (1) \textbf{Diverse Data Samples.} The dataset must include a variety of real-world scenes and object categories to provide sufficient and diverse samples for subsequent language annotations and support multi-object labeling. (2) \textbf{RGB-Thermal Spatial Alignment.} Each video frame must contain a pair of pixel-level aligned RGB and thermal infrare (T) images to facilitate effective information fusion across the RGB-T modalities. (3) \textbf{Continuous Video Sequences.} The original dataset should provide complete video streams containing continuous frames, in which the full motion trajectories of the targets can be observed throughout the frame sequence. After extensive evaluation across multiple datasets, we selected two public datasets, LasHeR \cite{li2021lasher} and VTUAV \cite{zhang2022visible}, that satisfy the aforementioned three criteria as the foundation of our dataset.

\begin{figure*}[!t]
    \setlength{\abovecaptionskip}{0.6mm}  
    \setlength{\belowcaptionskip}{-2.4mm}  
    \centering
    \includegraphics[width=\textwidth]{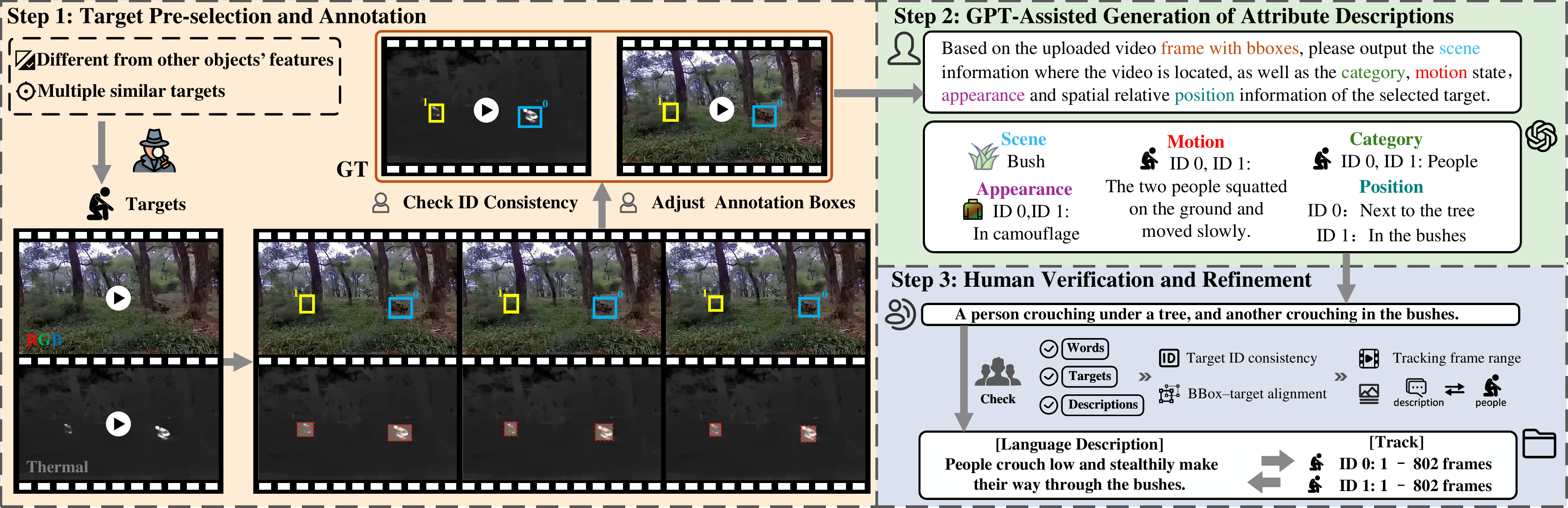} 
    \caption{\textbf{Data annotation process.} The annotation process consists of three steps: (1) Target Pre-selection and Annotation; (2) GPT-Assisted Generation of Attribute Descriptions; (3) Human Verification and Refinement. First, visually similar targets are selected, and bounding boxes are annotated across the entire sequence. Then, annotated key frames and task instructions are provided to GPT to analyze the target’s category, scene, appearance, motion, and spatial position. Finally, GPT-generated attributes are integrated into initial descriptions, which is refined through multiple annotators' review to produce the final language descriptions.}
    \label{fig:annotation process}
\end{figure*}

\subsection{Dataset Annotation}
\label{subsubsec:Dataset Annotation}

\quad To construct high-quality RGB-T modality data for referring multi-object tracking trajectories, we construct RefRT dataset based on two fundamental datasets using a GPT \cite{achiam2023gpt} assisted manual annotation strategy. The annotation process is depicted in Fig.~\ref{fig:annotation process}. The detailed workflow is as follows:

\noindent \textbf{Step\,1:\,Target Pre-selection and Annotation.} First, the annotator examines the aligned RGB–T video sequence and pre-selects several targets that share similar characteristics but are distinguishable from other objects. The annotator then labels bounding boxes for these targets across all video frames. During this process, the annotator ensures the accuracy of the bounding boxes and verifies the consistency of target IDs throughout the sequence.

\noindent \textbf{Step\,2:\,GPT-assisted Generation of Attribute Descriptions.} We feed GPT with RGB–T aligned annotated frames containing bounding boxes and target IDs, along with structured instructions specifying the attributes to be analyzed. GPT then generates key attributes such as target category, appearance features, spatial location, and other visual cues.

\noindent \textbf{Step\,3:\,Human Verification and Refinement.} First, an annotator organizes the attributes generated by GPT into an initial language description. Then, multiple reviewers verify the spelling, the consistency and validity of the description, and the accuracy of bounding boxes, as well as the correctness of target IDs and the alignment of bounding boxes with their corresponding targets. Finally, they check the tracking frame ranges of the targets and ensure that the described targets match the language description. After this review process, we obtain the final language description, along with the target IDs and the corresponding video frame ranges for each target matching the description.

\subsection{Dataset Split}
\label{subsubsec:Dataset Split}
\quad To ensure that the model has sufficient data to learn the relationships between RGB and thermal modalities during training while retaining enough unseen samples for evaluation. We split all video sequences in the RefRT dataset into training and test sets at a ratio of 6:4. The language descriptions and annotated targets associated with each video sequence are assigned to the corresponding subset along with the sequence. This results in a training set containing 42 videos and a test set containing 30 videos.

\begin{figure*}[t]
    \setlength{\abovecaptionskip}{2.6mm}  
    \setlength{\belowcaptionskip}{-2.4mm} 
     \centering
    \begin{subfigure}{0.47\linewidth}
      \includegraphics[width=1.0\linewidth]{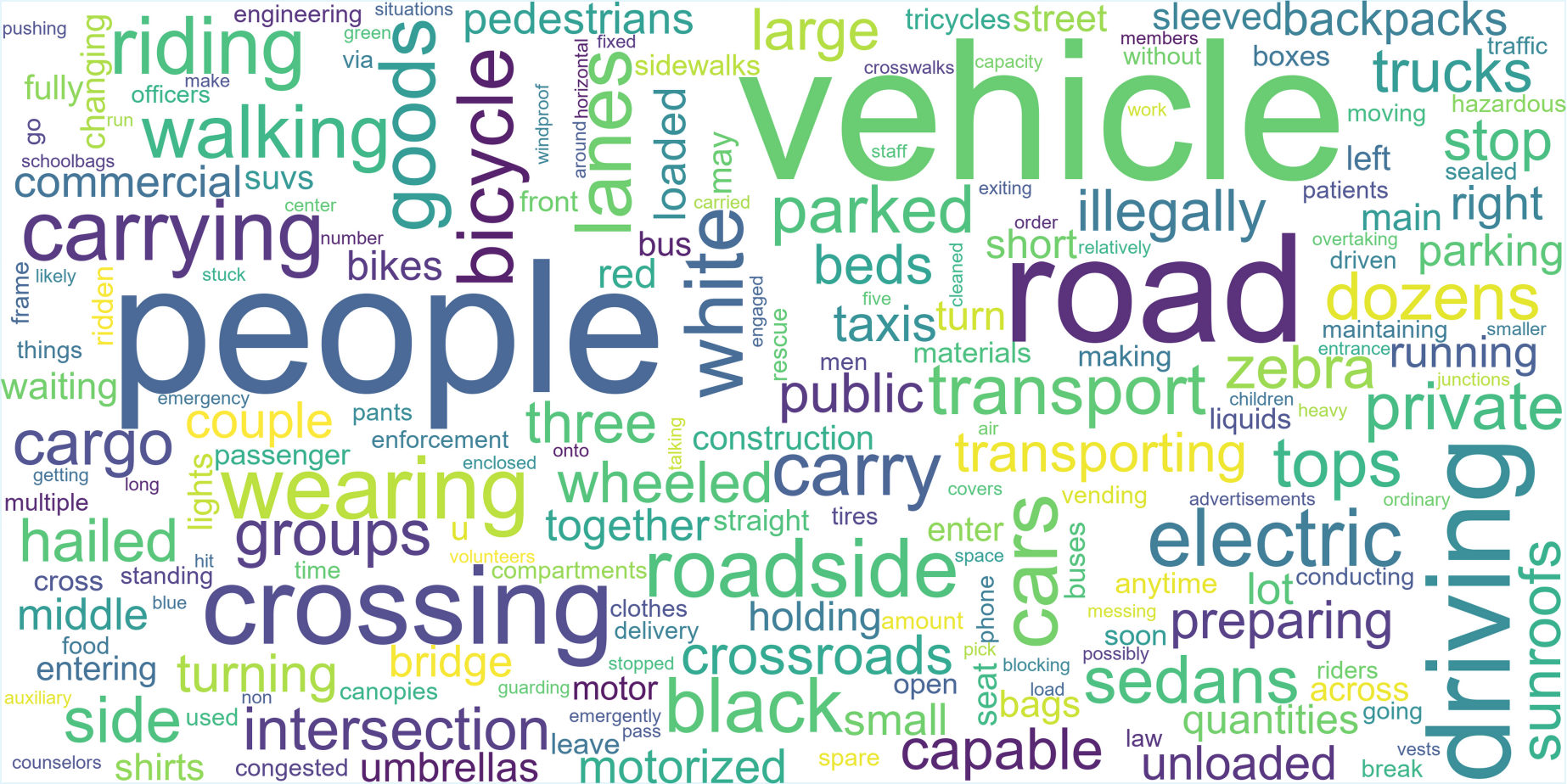}
      \caption{}
      \label{fig:a.}
    \end{subfigure}
    \hspace*{0.002\linewidth}
    \begin{subfigure}{0.24\linewidth} 
      \centering
      \includegraphics[width=1.0\linewidth]{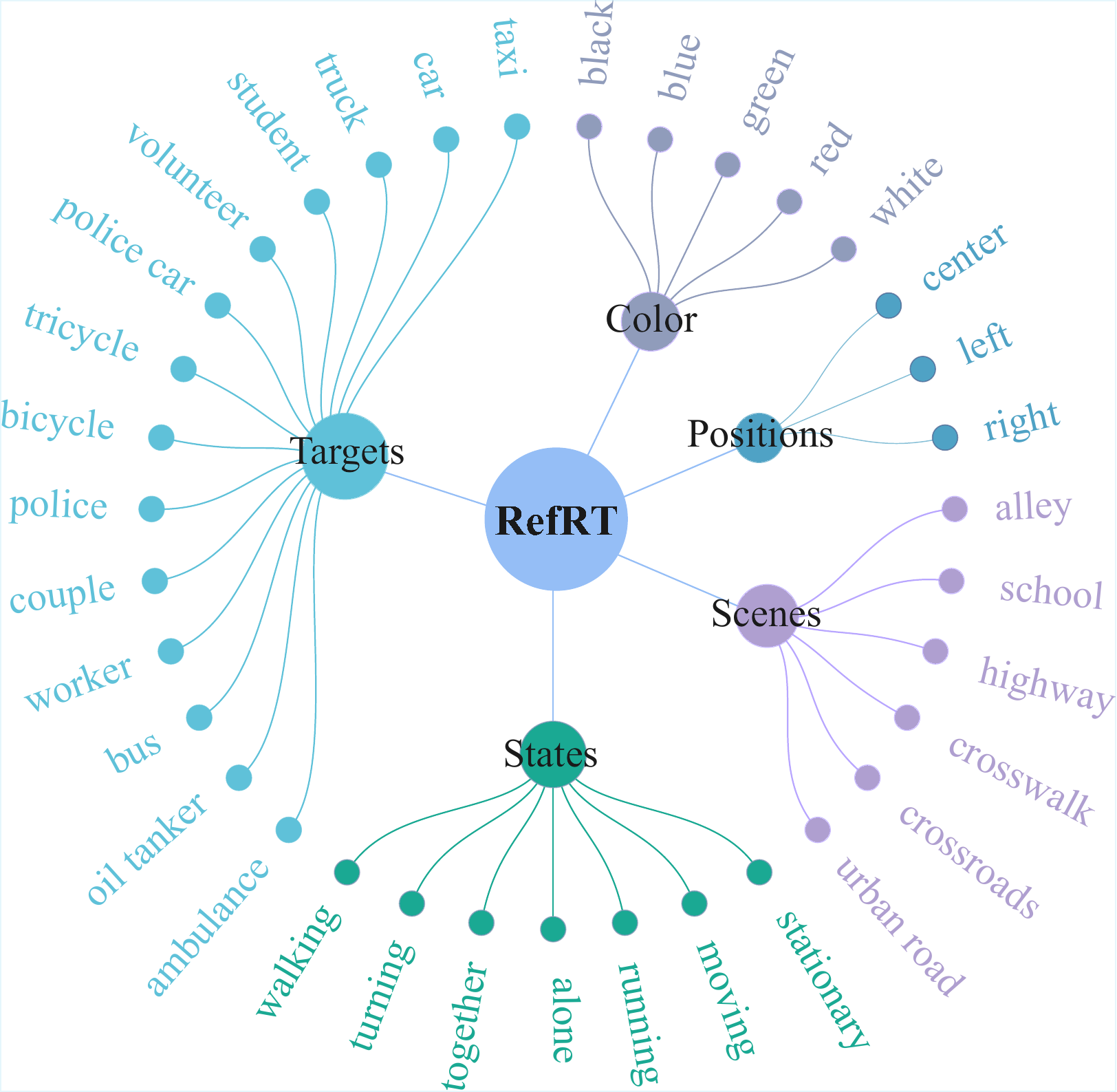}
      \caption{}
      \label{fig:c}
    \end{subfigure}
    \begin{subfigure}{0.25\linewidth} 
      \centering
      \includegraphics[width=1.0\linewidth]{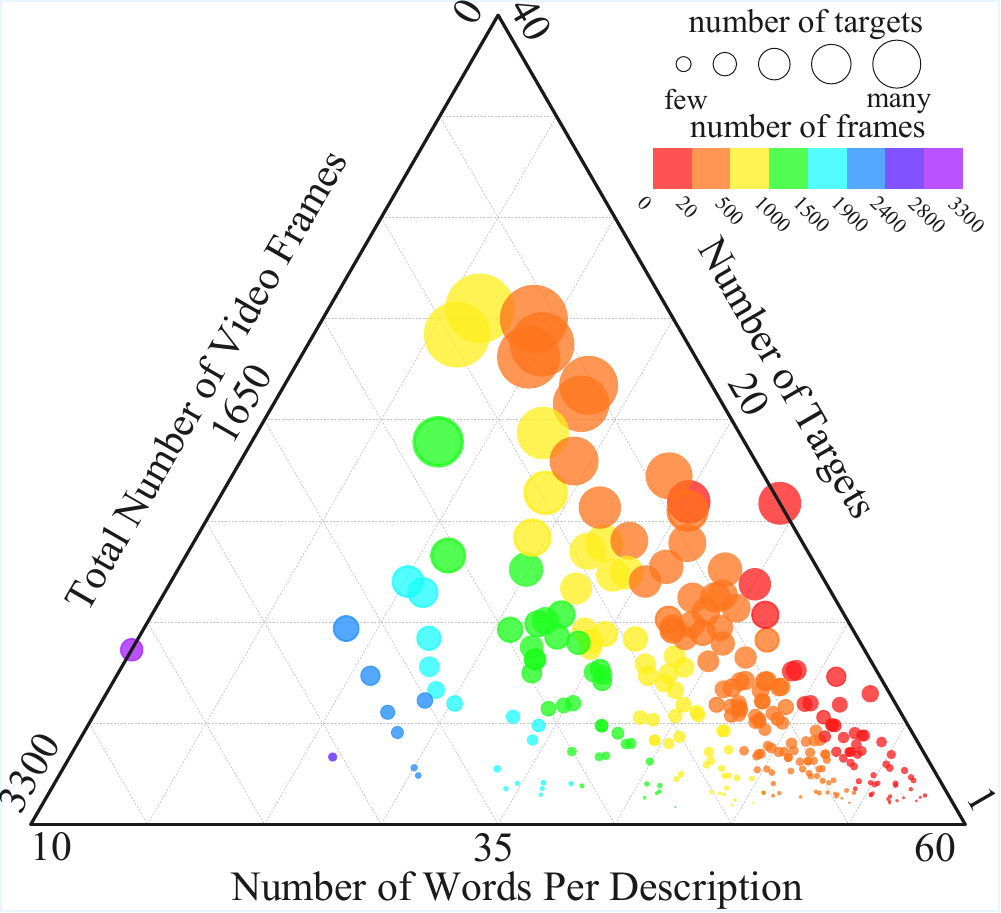}
      \caption{}
      \label{fig:b}
    \end{subfigure}
    \caption{\textbf{Visualization results on the RefRT dataset.} (a) The word cloud of the RefRT dataset contains a rich set of keywords. (b) The RefRT dataset covers diverse scenes, targets and attributes. (c) After normalizing the three-dimensional data of the language descriptions in the RefRT dataset, the broad distribution characteristics of the dataset are revealed.}
    \label{fig:RT_rmots_three}
\end{figure*}

\subsection{Dataset Statistics} 
\label{subsubsec:Dataset Statistics}
\quad The RefRT dataset contains 72 scenes, 166,147 RGB–Thermal–Language triplets, 388 language descriptions, and 1,250 targets with language annotations. It features pixel-level RGB–thermal alignment, covers both standard and low visibility scenarios, and includes diverse targets such as pedestrians and vehicles. Below, we present a comprehensive analysis of the dataset.

\noindent \noindent \textbf{Word Cloud.} We perform a word cloud analysis of the RefRT dataset, and the results are shown in Fig.~\ref{fig:RT_rmots_three} (a). RefRT encompasses rich semantic information and provides detailed language descriptions through multi-dimensional descriptions, including category attributes, behavioral characteristics, and spatial locations. This highlights the diversity of the RefRT dataset.

\noindent \textbf{RefRT Language Analysis.} As shown in Fig.~\ref{fig:RT_rmots_three} (b), we conduct a detailed analysis of the language descriptions in the RefRT dataset. These descriptions cover common scenes such as urban areas, roads, and schools, demonstrating the dataset’s strong real-world relevance. Moreover, to align with the RT-RMOT task, the language descriptions focus on targets capable of emitting thermal radiation, such as pedestrians, cars, and electric vehicles. The associated language descriptions capture behavioral cues, scene attributes, and inter-object interactions, further highlighting the dataset’s rich semantic complexity.

\noindent \textbf{Data Distribution Characteristics.} Each data point in Fig.~\ref{fig:RT_rmots_three} (c) represents a semantic description, where the three axes denote word count, tracking frames, and target numbers. The figure shows that the RefRT dataset presents a broad and balanced distribution across these dimensions. By including descriptions of varying lengths and complexities, RefRT meets model learning requirements. Variations in target instances challenge the model’s tracking ability, while different frame spans evaluate its generalization under diverse temporal dynamics.

\begin{figure*}[t]
    \setlength{\abovecaptionskip}{0.8mm}  
    \setlength{\belowcaptionskip}{-3.4mm}  
    \centering
    \includegraphics[width=\textwidth]{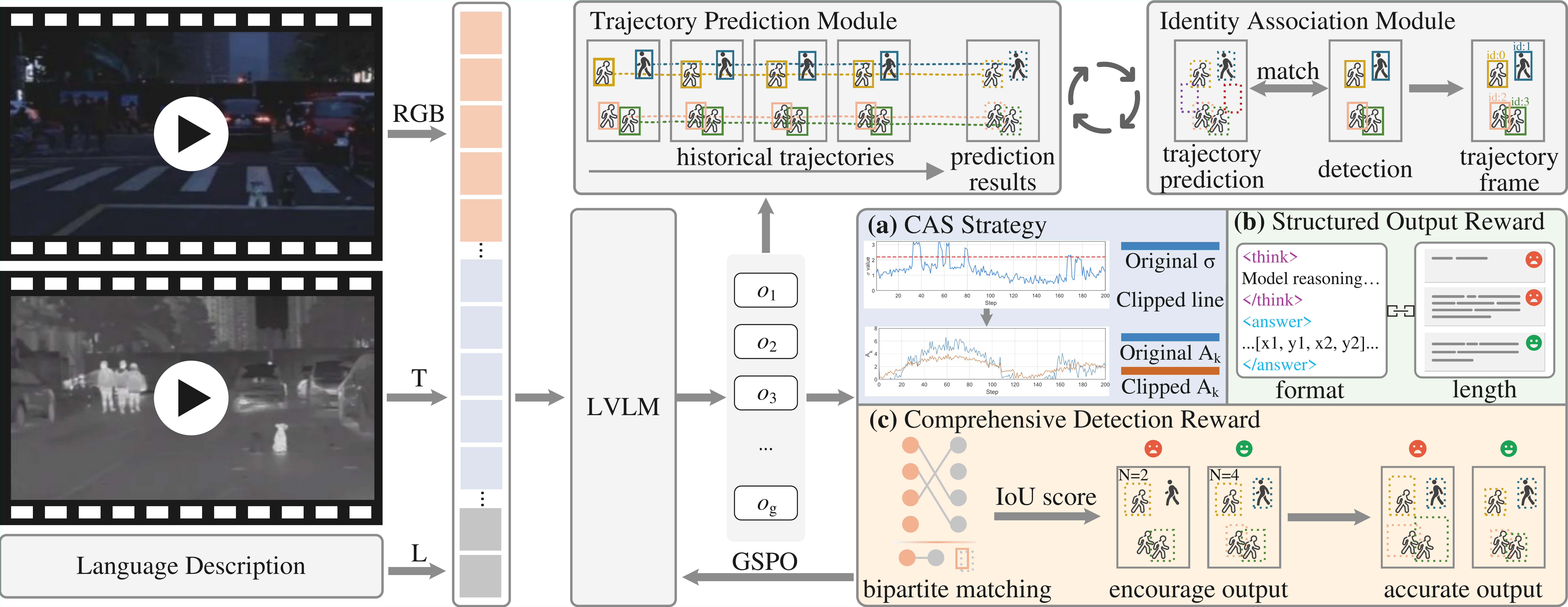}
    \caption{\textbf{Overall pipeline of RTrack.} The RTrack framework consists of three modules: (1) Large-model perception module: performs inference detection of the target in the current frame based on the language description. (2) Trajectory Prediction Module: predicts the target box in the current frame based on the historical trajectory. (3) Identity Association Module: matches the trajectory box and the detection box to generate the target ID. Among them, the GSPO algorithm includes three aspects: (a) CAS strategy: constrains relative rewards to avoid gradient explosion. (b) Structured output reward: requires the model to format the output and limits the output length. (c) Comprehensive detection reward: completes the output encouragement and accurate output based on the IoU value.}
    \label{fig:r2gbt_framework}
\end{figure*}

\section{Methodology}
\label{sec:Methodology}
\quad To address this challenge of the RT-RMOT tasks, we propose RTrack, which leverages the perception capability of Multimodal Large Language Models (MLLMs) to detect and associate multiple targets.

\subsection{Framework}
\label{sec:Framework}
\quad We propose \textbf{RTrack}, a framework for the RGB-Thermal Referring Multi-Object Tracking (RT-RMOT) task. It leverages the strengths of \textbf{Multi-modal Large Language Models (MLLMs)} in RGBT modal fusion and semantic reasoning. Unlike traditional RMOT models that only process RGB images, RTrack can accurately locate and continuously track referred targets in RGBT scenarios. As illustrated in Fig.~\ref{fig:r2gbt_framework}, RTrack consists of three main components: \textbf{Large-model Perception Module}, \textbf{Trajectory Prediction Module}, and \textbf{Identity Association Module}.

\textbf{MLLM-based Perception Module.} We introduce a perception mechanism based on a Multimodal Large Language Model (MLLM), enabling the model to simultaneously achieve RGBT feature fusion and language understanding, thereby realizing dual-modality object perception and localization. Specifically, given an aligned pair of RGBT images $\{I^{\text{V}}, I^{\text{T}}\}$ and a language description $L$, the MLLM sequentially performs visual encoding $E_v$, text encoding $E_l$, modality fusion \textit{Fuse}, cross-modal attention \textit{CrossAttn}, and visual instruction reasoning \textit{MLLM}, ultimately obtaining the spatial location of the target specified by the language description in the RGBT image. 

\textbf{Trajectory Prediction Module.} This module dynamically model historical trajectory sequences and analyze motion trends, thereby generating prior predictions of target positions in the current frame. It employs a Kalman Filter \cite{kalman1960new} to perform linear modeling of the targets’ states, including position and velocity, and achieves continuous tracking through a prediction–update loop. In each frame, the filter first predicts the targets’ positions based on the previous state, then updates the state using the detection results from the Perception Module. 

\textbf{Identity Association Module.} This module maintains target identity consistency across consecutive frames and ensure trajectory continuity. Specifically, the system first generates candidate targets for the current frame based on the output of the Perception Module, and utilizes predicted bboxes from the Trajectory Prediction Module to construct an IoU-based cost matrix between the predicted and detection bboxes. On this basis, the Hungarian Matching Algorithm \cite{kuhn1955hungarian} is applied to achieve global optimal matching, thereby determining the identity correspondences between detection bboxes and historical trajectories. For successfully matched trajectories, the system updates their states using current detections; unmatched detections are initialized as new trajectories, while unmatched trajectories that exceed their predefined lifespan are considered lost and consequently terminated from tracking. The algorithm of the identity association module is illustrated in \cref{alg:identity_association}.

\subsection{RL Fine-tuning based on the GSPO algorithm}
\label{sec:RL Fine-tuning based on the GSPO algorithm}

\quad To further enhance cross-modal fusion capability, language understanding, and tracking performance of the RTrack framework on the RefRT dataset, we employ the Group Sequence Policy Optimization (GSPO) \cite{zheng2025group} algorithm to fine-tune the model using reinforcement learning. To enhance the stability of training and cross-modal detection capability, we make improvements in the GSPO framework from both the policy and reward dimensions.

\begin{algorithm}[t]
\setlength{\abovecaptionskip}{0.8mm}
\setlength{\belowcaptionskip}{-2.4mm}
\caption{\textbf{Identity Association Module}}
\label{alg:identity_association}

\textbf{Input}: Detection boxes $\mathcal{D}$; predicted trajectories $\mathcal{T}^{pred}$.\\
\textbf{Parameter}:  Calculation operation $\mathcal{C}()$; Hungarian algorithm $\mathcal{H}()$; IoU cost matrix $M_{iou}$; matched/unmatched sets $\{\mathcal{T}^{m}, \mathcal{T}^{u}, \mathcal{D}^{m}, \mathcal{D}^{u}\}$; maximum lifespan $\delta_{max}$.

\begin{algorithmic}[1]
\Statex \textbf{/* IoU-based matching */}
\State $M_{iou} \gets \mathcal{C}(\mathcal{D}, \mathcal{T}^{pred})$
\State $(\mathcal{T}^{m}, \mathcal{T}^{u}, \mathcal{D}^{m}, \mathcal{D}^{u}) \gets \mathcal{H}(M_{iou})$

\Statex \textbf{/* Update matched trajectories */}
\ForAll{$(\mathcal{T}^{m}_{i}, \mathcal{D}^{m}_{i})$}
    \State Update $\mathcal{T}^{m}_{i}$ with $\mathcal{D}^{m}_{i}$ at frame $F_c$
    \State $\mathcal{T}^{m}_{i}.\text{status} \gets$ \textit{active}; reset missing count
\EndFor

\Statex \textbf{/* Initialize new trajectories */}
\ForAll{$\mathcal{D}^{u}_{i}$}
    \State $\mathcal{T}^{new}_{i} \gets \text{Init}(\mathcal{D}^{u}_{i}, F_c)$
    \State $\mathcal{T}^{new}_{i}.\text{status} \gets$ \textit{active}; add to $\mathcal{T}$
\EndFor

\Statex \textbf{/* Handle unmatched trajectories */}
\ForAll{$\mathcal{T}^{u}_{i}$}
    \State $\mathcal{T}^{u}_{i}.\text{missing\_count} \gets \mathcal{T}^{u}_{i}.\text{missing\_count} + 1$
    \If{$\mathcal{T}^{u}_{i}.\text{missing\_count} > \delta_{max}$}
        \State $\mathcal{T}^{u}_{i}.\text{status} \gets$ \textit{terminated}
    \Else
        \State $\mathcal{T}^{u}_{i}.\text{status} \gets$ \textit{temporary}
    \EndIf
\EndFor

\State \Return $\mathcal{T}$
\end{algorithmic}
\end{algorithm}

Group Sequence Policy Optimization (GSPO) is a sequence-level policy optimization algorithm for the reinforcement learning stage of large language models. It treats the entire generated sequence as the optimization unit, aligning the granularity of policy updates with reward evaluation. For each input, the model generates multiple candidate sequences, computes their average reward as a baseline, and updates the policy by relative rewards. Unlike token-level methods (e.g., GRPO \cite{shao2024deepseekmath}), GSPO optimizes at the sequence level with an importance ratio and length-normalized clipping mechanism, effectively reducing training noise and instability. Its objective is formulated as:
\begin{equation}
\scriptsize
\begin{aligned}
\mathcal{J}_{\text{GSPO}}(\theta)
&=
\mathbb{E}_{x\!\sim\!\mathcal{D},\,\{y_k\}\!\sim\!\pi_{\theta_{\text{old}}}(\cdot|x)}
\!\left[
\frac{1}{G}\!\sum_{k=1}^{G}
\min(s_1A_k,s_2A_k)
\right] \\[-2pt]
&\quad
-\beta_{\mathrm{KL}}\,
\mathbb{E}_{x\!\sim\!\mathcal{D}}\!\left[
D_{\mathrm{KL}}\!\big(
\pi_\theta(\cdot|x)\,\|\,\pi_{\theta_{\text{old}}}(\cdot|x)
\big)
\right]
\end{aligned}
\label{eq:gspo_kl}
\end{equation}

\begin{equation}
\begin{aligned}
s_1(\theta)
&=
\left(
\frac{\pi_\theta(y_k\,|\,x)}
{\pi_{\theta_{\text{old}}}(y_k\,|\,x)}
\right)^{\frac{1}{|y_k|}}
\end{aligned}
\end{equation}
\begin{equation}
\begin{aligned}
s_2 = \text{clip}\left( s_1(\theta), 1+\varepsilon, 1-\varepsilon \right)
\end{aligned}
\label{eq:gspo_combined}
\end{equation}
Where \( \varepsilon \) and \( \beta_{\text{KL}} \) are hyperparameters, and \( A_k \) denotes the relative advantage of the \( k \)-th generated sequence.

\textbf{Clipped Advantage Scaling (CAS) Strategy.} To mitigate gradient explosion that may arise when the relative rewards are standardized, leading to abnormal amplification of the advantage term, especially when the reward distribution within the group becomes concentrated, we constructed a clipped, normalized advantage based on the CAS strategy. Specifically, given a set of scalar rewards \( \{r_k\}_{k=1}^G \) with their group mean \( \mu \) and standard deviation \( \sigma \), the formula for calculating the new normalized advantage is as follows:
\begin{equation}
A_k \;=\; (r_k - \mu) \cdot \mathrm{clip}\!\left(\frac{1}{\sigma},\, 0,\, Scale_{\max}\right) 
\end{equation}
where $Scale_{\max}$ controls the maximum global amplification factor, indicating the maximal multiplier of the advantage during advantage normalization. 

\textbf{Rule-based Reward Function.} 
To effectively guide the model in generating structured and high-precision outputs, 
we comprehensively consider the reward rules of the task and design a rule-based composite reward mechanism consisting of two weighted parts, namely the structured output reward 
$R_{\text{str}}$ and the comprehensive detection reward 
$R_{\text{ctr}}$.

\textit{\textbf{Structured Output Reward.}} To standardize the output and balance exploration and exploitation, we introduce a structured output reward. The structured output reward consists of two parts: the format reward and the length reward, which are weighted and summed.

Specifically, the model’s reasoning process must be enclosed within the \texttt{<think}
\texttt{></think>} tags, and the final conclusion within the \texttt{<answer></answer>} tags. To ensure detectable outputs, the conclusion must include at least one coordinate set in the format \texttt{[x1,y1,x2,y2]}. If the output fully meets these formatting rules, \( R_{\text{format}}=1 \); otherwise, \( R_{\text{format}}=0 \). In addition, a sine-window--smoothed length reward \( R_{\text{len}}(L) \in [0, 1] \) is applied to regularize response length.
\begin{equation}
s(x) = \min \left( \max \left( \sin^2 \left(\frac{\pi}{2}x\right), 0 \right), 1 \right)
\end{equation}
\begin{equation}
R_{\text{len}}(L) = s \left(\frac{L - L_{\min}}{L_{\text{low}} - L_{\min}}\right) \cdot s \left(\frac{L_{\max} - L}{L_{\max} - L_{\text{high}}}\right)
\end{equation}
where \( L \) means the response length and is obtained by summing over the completion mask.

\textit{\textbf{Comprehensive Detection Reward.}} To balance the comprehensiveness and accuracy of multi-object detection, we propose a reward based on the Intersection over Union (IoU) between detected and ground-truth boxes. The reward comprises two components: \textbf{Output Encouragement Reward} and \textbf{Precision Detection Reward}. Given the detection set \( B_{\text{det}} \) and ground-truth set \( B_{\text{gt}} \), we compute the IoU matrix after mapping all detected boxes to the same coordinate space as \( B_{\text{gt}} \). A cost matrix is then constructed based on the IoU values, and the Hungarian algorithm is applied for optimal one-to-one matching. The weighted sum of matched IoUs, denoted as \( IoU_{\text{score}} \), serves as the basis for subsequent reward computation.

To enhance the model’s exploration ability and encourage it to identify as many potential targets related to the language instructions as possible, we designed \textbf{Output Encouragement Reward} that encourages the model to output more predictions. As the number of valid matched boxes detected by the model increases, its reward also increases, thereby guiding the model to identify and cover more target areas, striving to ensure that all ground truth targets are covered. This reward can be defined as:
\begin{equation}
R_{OER}(B_{det}, B_{gt}) = \alpha \cdot MatchedGT + \beta \cdot IoU_{score}
\label{eq:oer}
\end{equation}
where \textit{MatchedGT} denotes the number of valid ground-truth boxes that are successfully matched with predictions.

In the later stages of training, to further improve the model’s detection accuracy, we introduce the \textbf{Precision Detection Reward}. This reward not only requires the predicted bounding boxes to be spatially close to the ground truth boxes (i.e., having a high IoU), but also prevents the model from repeatedly outputting excessive predictions near high IoU regions to “boost” the reward. Additionally, this mechanism prevents the model from neglecting samples that are close to the ground truth but have slightly lower IoU values due to overemphasis on precision. The Precision Detection Reward can be defined as:
\begin{equation}
R_{PDR}(B_{\text{det}}, B_{\text{gt}}) =
\frac{IoU_{score}}{(N_{\text{det}})^{\gamma}} +
\frac{\lambda \cdot MatchedGT}{N_{\text{gt}}}
\end{equation}
where \( IoU_{\text{score}} \) and \textit{MatchedGT} have the same meanings as defined in Eq.~(\ref{eq:oer}). Here, $N_{\text{det}}$ and $N_{\text{gt}}$ denote the numbers of predicted and ground-truth boxes.

\section{Experiments}
\label{sec:Experiments}
\subsection{Implementation Details}
\quad Our RTrack framework is built upon a Multi-modal Large Language Model. Specifically, we use Qwen2.5-VL-3B as the MLLM baseline. All experiments are conducted on an NVIDIA RTX 3090 GPU. The RL phase uses LORA fine-tuning with an initial learning rate of \(1 \times 10^{-5}\) with 4 rollouts by default. We temporally and uniformly sample 10\% of the frames from the training set. Some key parameter settings are shown in the supplementary material.

\begin{table*}[t]
\setlength{\abovecaptionskip}{2mm}   
\centering
\caption{Performance of many methods on the RefRT test set. ↑ indicates that higher scores are better. The best results are marked in \textbf{bold}.}
\label{tab:Qualitative Results.}

\vspace{-5pt}
\resizebox{1.0\linewidth}{!}{
    \begin{tabular}{llccccccccccccc}
    \toprule
    \textbf{Modality} & \textbf{Method} & \textbf{Venue} & HOTA↑ & DetA↑ & AssA↑ & DetRe↑ & DetPr↑ & AssRe↑ & AssPr↑ & LocA↑ \\
    \midrule
    \multirow{5}{*}{\textbf{RGB}} 
     & TransRMOT \cite{wu2023referring} & CVPR 2023 & 8.69 & 2.57 & 29.96 & 3.01 & 14.46 & 30.73 & 85.49 & \textbf{79.63} \\
     & TempRMOT \cite{zhang2024bootstrapping} & ArXiv 2024 & 8.19 & 1.86 & 36.23 & 2.04 & 16.68 & 39.28 & 75.39 & 77.48 \\
     & \underline{CRTracker} \cite{chen2025cross} & AAAI2025 & 9.30 & 2.37 & 37.01 & 3.81 & 5.83 & 40.10 & 67.48 & 73.25 \\
     & YOLOX+ByteTrack+\underline{iKUN} \cite{du2024ikun} & CVPR 2024 & 2.32 & 0.29 & 19.86 & 0.29 & 12.71 & 21.18 & 61.45 & 69.70 \\
     & Qwen2.5-VL-3B~\cite{bai2025qwen2} & ArXiv 2025 & 2.09 & 0.93 & 5.28 & 0.97 & 17.14 & 5.40 & \textbf{87.46} & 76.69 \\
    \midrule
    \multirow{6}{*}{\textbf{RGBT}} 
     & \underline{DeformCAT} \cite{hu2025deformle}+SORT+iKUN & IEEE TMM & 2.03 & 0.41 & 11.25 & 0.77 & 0.87 & 12.07 & 47.65 & 62.61 \\
     & \underline{Unismot}+iKUN \cite{ZHANG2025110984} & PR 2025 & 1.95 & 0.29 & 14.34 & 0.31 & 3.98 & 15.41 & 65.48 & 70.86 \\
     & \underline{PFTrack}+iKUN \cite{zhu2025visible} & PR 2025 & 8.55 & 1.66 & \textbf{45.92} & 2.40 & 5.05 & \textbf{49.15} & 73.96 & 76.31 \\
     & \underline{MCTrack}+iKUN \cite{ma2025multi} & TCSVT 2025 & 4.71 & 1.22 & 18.91 & 1.51 & 5.73 & 19.83 & 71.17 & 68.95 \\
     & Qwen2.5-VL-3B(baseline) & ArXiv 2025 & 4.98 & 2.59 & 10.19 & 3.05 & 14.29 & 10.65 & 83.40 & 75.52 \\
     & \textbf{RTrack} & \textbf{Ours} & \textbf{15.53} & \textbf{12.39} & 20.79 & \textbf{20.15} & \textbf{22.78} & 22.02 & 81.99 & 75.53 \\
    \bottomrule
    \end{tabular}
}
\vspace{-5pt}
\end{table*}

\subsection{Metrics}
\quad We evaluate our tracking framework following the RMOT community’s evaluation protocol, using HOTA for comprehensive assessment of detection and tracking performance. In addition, DetA, DetPr, and DetRe measure detection quality. AssA, AssPr, and AssRe assess tracking consistency. LocA evaluates bounding box accuracy.

\subsection{Quantitative Results}
\quad To evaluate the performance of the proposed RTrack framework on the RefRT dataset, we conduct a comprehensive set of comparative experiments. We select four representative RMOT methods: the end-to-end approaches TransRMOT \cite{wu2023referring}, TempRMOT \cite{zhang2024bootstrapping}, and CRTracker \cite{chen2025cross}, as well as the two-stage method iKUN \cite{du2024ikun}. Since iKUN is a two-stage method that requires an external detector and tracker, we adopt YOLOX \cite{ge2021yolox} + ByteTrack \cite{zhang2022bytetrack} as its tracking module, considering the diverse categories in the RefRT dataset. In addition, these methods cannot directly process thermal (T) images. To ensure a fair comparison, we only evaluate their performance on the RGB modality and compared it with Qwen2.5-VL-3B \cite{bai2025qwen2} (RGB input). To enable comparison under the RGBT modality, we design a solution that supports RGBT input, namely a RGBT tracker + iKUN combination, and compare it with the RTrack framework (RGBT input). Here, DeformCAT serves as the RGBT detector, while Unismot, PFTrack, and MCTrack are the RGBT trackers. As shown in \cref{tab:Qualitative Results.}, in the RGB modality, the Qwen2.5-VL baseline performs slightly worse than other RMOT-specific methods, which is consistent with the fact that the latter are specially designed for the RMOT task, whereas Qwen2.5-VL is a more general model with stronger generalization capabilities. However, in the RGBT modality, Qwen2.5-VL significantly outperforms the RGBT tracker + iKUN combination, demonstrating its superior ability in handling RGBT input and language understanding.

\subsection{Ablation Study}
\quad To further verify the effectiveness of the proposed task, framework, and design mechanism, we conduct three groups of ablation experiments: multimodal input, MLLMs and the design mechanism, as shown in \cref{tab:Experimental results of multimodal input.} and \cref{tab:ablation study}.

\textbf{Validation of Multimodal Input Effectiveness.} We compare the performance of the RTrack framework under different input modalities. Specifically, we conduct experiments with both RGB and RGB–Thermal (RGBT) inputs under zero-shot settings and after RL fine-tuning. As shown in \cref{tab:Experimental results of multimodal input.}, the proposed training strategy achieves significant performance improvements under both RGB and RGBT input settings. Specifically, with RGB inputs, the trained model improves HOTA by 10.4\% over the untrained version and yields consistent improvements across the five key metrics: DetA, AssA, DetRe, DetPr, and AssRe. With RGBT inputs, the trained model improves HOTA by 10.55\% and similarly achieves substantial gains on these five metrics. Furthermore, comparing the trained models reveals that using RGBT as input leads to superior overall performance compared to RGB, achieving a 3.04\% higher HOTA score and outperforming RGB on five detection and association metrics. These findings collectively demonstrate the importance of leveraging RGBT multimodal information to enhance the performance of the RTrack framework.
\begin{table*}[t]
\setlength{\abovecaptionskip}{2mm}   
\centering
\caption{Results of more experiments on the RefRT test set.↑ indicates that higher scores are better. The best results are marked in \textbf{bold}.}
\label{tab:Experimental results of multimodal input.}
\vspace{-5pt}
\resizebox{1.0\linewidth}{!}{
    \begin{tabular}{llccccccccccccc}
    \toprule
    \textbf{Modality} & \textbf{Method} & \textbf{Size} & \textbf{RL Fine-tuning} & HOTA↑ & DetA↑ & AssA↑ & DetRe↑ & DetPr↑ & AssRe↑ & AssPr↑ & LocA↑ \\
    \midrule
    \multirow{2}{*}{\textbf{RGB}} 
     & Qwen2.5-VL~\cite{bai2025qwen2} & 3B & \ding{55} & 2.09 & 0.93 & 5.28 & 0.97 & 17.14 & 5.40 & \textbf{87.46} & \textbf{76.69} \\
     & RTrack & 3B & \ding{51} & 12.49 & 11.05 & 15.25 & 16.14 & \textbf{23.95} & 16.18 & 79.39 & 73.90 \\
    \midrule
    \multirow{2}{*}{\textbf{RGBT}} 
     & Qwen2.5-VL~\cite{bai2025qwen2} & 3B & \ding{55} & 4.98 & 2.59 & 10.19 & 3.05 & 14.29 & 10.65 & 83.40 & 75.52 \\
     & RTrack & 3B & \ding{51} & \textbf{15.53} & \textbf{12.39} & \textbf{20.79} & \textbf{20.15} & 22.78 & \textbf{22.02} & 81.99 & 75.53 \\
    \bottomrule
    \end{tabular}
}
\vspace{-7pt}
\end{table*}

\begin{table*}[t]
\centering
\caption{\textbf{Ablation study results} of RTrack on the RefRT test set. Module A represent the untrained results, while Module B reflects the results after training. 
↑ indicates higher is better. Best results are in \textbf{bold}.}
\label{tab:ablation study}
\vspace{-5pt}
\setlength{\tabcolsep}{3pt}
\scriptsize
\resizebox{1.0\linewidth}{!}{
\begin{tabular}{llcccccccccccccc}
\toprule
\textbf{Module} & \textbf{Method} & \textbf{Size} & \textbf{RL Fine-tuning} & HOTA↑ & DetA↑ & AssA↑ & DetRe↑ & DetPr↑ & AssRe↑ & AssPr↑ & LocA↑ \\
\midrule
\multirow{4}{*}{\textbf{A. MLLMs}} 
& LLaVA~\cite{liu2023visual} & 7B & \ding{55} & 1.13 & 0.29 & 5.32 & 8.84 & 0.28 & 7.32 & 19.47 & 57.69 \\
& LLaVA-NeXT~\cite{liu2024llavanext} & 8B & \ding{55} & 2.99 & 0.84 & 13.35 & 5.70 & 0.96 & 17.84 & 38.32 & 60.17 \\
& InternVL2~\cite{chen2024internvl} & 8B & \ding{55} & 0.82 & 0.18 & 4.77 & 10.77 & 0.18 & 5.71 & 31.96 & 60.10 \\
& Qwen2.5-VL~\cite{bai2025qwen2} & 3B & \ding{55} & 4.98 & 2.59 & 10.19 & 3.05 & 14.29 & 10.65 & 83.40 & 75.52 \\
\midrule
\multirow{4}{*}{\textbf{B. Training Strategy}} 
& w/o CAS & 3B & \ding{51} & 5.27 & 3.93 & 7.37 & 4.32 & \textbf{28.11} & 7.43 & \textbf{90.32} & \textbf{78.63} \\
& w/o Struct. Reward & 3B & \ding{51} & 11.24 & 9.47 & 14.25 & 16.81 & 17.10 & 14.79 & 86.08 & 77.20 \\
& w/o Det. Reward & 3B & \ding{51} & 11.16 & 7.82 & 17.42 & \textbf{28.44} & 9.50 & 18.33 & 82.64 & 76.34 \\
& \textbf{RTrack (Ours)} & 3B & \ding{51} & \textbf{15.53} & \textbf{12.39} & \textbf{20.79} & 20.15 & 22.78 & \textbf{22.02} & 81.99 & 75.53 \\
\bottomrule
\end{tabular}
}
\vspace{-15pt}
\end{table*}

\textbf{Analysis of Multi-modal Large Language Models.} Under the same RTrack framework, we replace different vision-language models (LLaVA, LLaVA-NeXT, InternVL and Qwen2.5-VL) and conduct zero-shot inference comparisons to evaluate the differences among large vision-language models in multimodal perception and instruction understanding. As shown in section A of \cref{tab:ablation study}, we observe that although Qwen2.5-VL uses the 3B model, which has the smallest number of parameters among the compared models, it still achieves the best performance. This confirms that choosing Qwen2.5-VL as the baseline for RTrack provides better modality fusion and detection capabilities.

\textbf{Analysis of Training Strategy Improvements.} During training, we perform an ablation study on the RTrack optimization mechanism to evaluate the contributions of the improved GSPO algorithm and the multidimensional reward design. Each component’s impact on training stability and performance is assessed by progressively removing it. As shown in section B of \cref{tab:ablation study}, the full model achieves the best detection performance, tracking performance, and multi-target matching accuracy. Specifically, the Cropping Advantage Scaling (CAS) strategy significantly improves the model’s tracking performance by mitigating gradient fluctuations caused by group normalization; the Structured Output Reward plays a key role in constraining the rationality of output structures and maintaining stable output lengths; the Comprehensive Detection Reward improves the accuracy and completeness of multi-target detection. Overall, the designed RL fine-tuning strategy enhances the model’s semantic understanding and referring multi-object tracking capabilities.

\subsection{Visualization}
\begin{figure*}[t]
    \setlength{\abovecaptionskip}{0.8mm}   
    \setlength{\belowcaptionskip}{-2.4mm}  
    \centering
    \includegraphics[width=\textwidth]{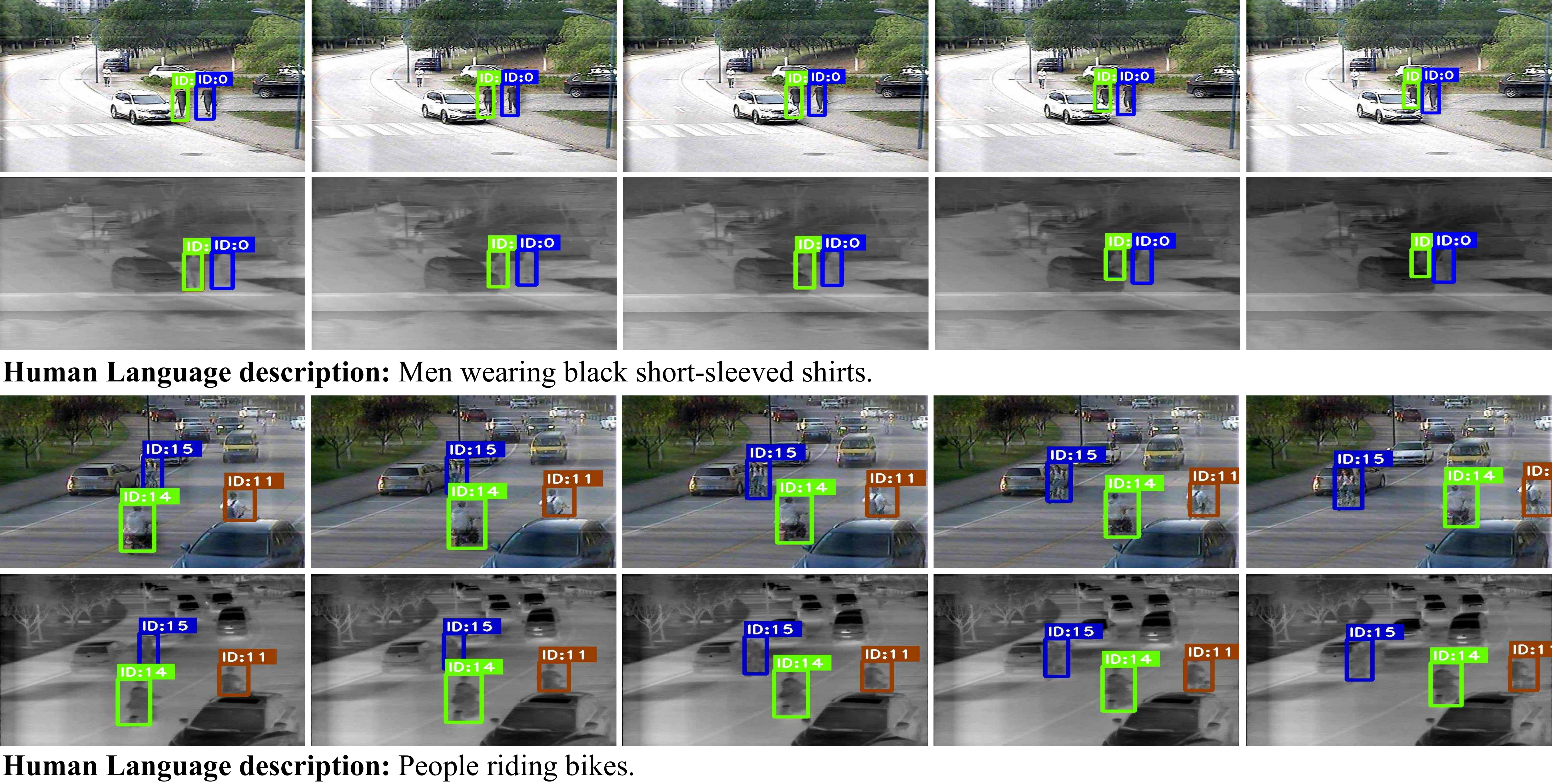}
    \caption{Qualitative zero-shot results of RTrack on the test set of RT-RMOT task.}
    \label{fig:Visualization}
\end{figure*}

\quad To further verify the effectiveness of the proposed RTrack framework, we conduct qualitative visualization analysis. As shown in \cref{fig:Visualization}, RTrack accurately detects and continuously tracks targets following the given language descriptions. By fusing complementary RGB and thermal infrared (T) information, the model maintains stable performance under challenging conditions such as low illumination, occlusion, and complex backgrounds. These results highlight the advantages of multimodal feature fusion in improving semantic understanding.

\section{Conclusion}
\label{sec:Conclusion}
\quad This work proposes a new task, \textbf{R}GB\textbf{T} \textbf{R}eferring \textbf{M}ulti-\textbf{O}bject \textbf{T}racking (\textbf{RT-RMOT}), which aims to overcome the challenges of tracking under low-visibility conditions such as nighttime and smoke, enabling all-day referring multi-object tracking. To advance research, we construct the first \textbf{RefRT} dataset, featuring diverse scenarios and rich language descriptions, providing a benchmark for language-guided multi-object tracking in challenging environments. In addition, we propose the \textbf{RTrack} framework, which leverages MLLMs to enhance modality fusion and incorporates a CAS strategy to stabilize gradients, along with a Structured Output Reward and Comprehensive Detection Reward to balance exploration and accuracy. Experiments on RefRT dataset demonstrate that RTrack achieves \textbf{state-of-the-art (SOTA)} performance, providing a strong baseline for future research.

\bibliographystyle{splncs04}
\bibliography{main}

@String(AAAI  = {AAAI})

@inproceedings{wu2023referring,
  title={Referring multi-object tracking},
  author={Wu, Dongming and Han, Wencheng and Wang, Tiancai and Dong, Xingping and Zhang, Xiangyu and Shen, Jianbing},
  booktitle={Proceedings of the IEEE/CVF conference on computer vision and pattern recognition},
  pages={14633--14642},
  year={2023}
}

@inproceedings{du2024ikun,
  title={ikun: Speak to trackers without retraining},
  author={Du, Yunhao and Lei, Cheng and Zhao, Zhicheng and Su, Fei},
  booktitle={Proceedings of the IEEE/CVF Conference on Computer Vision and Pattern Recognition},
  pages={19135--19144},
  year={2024}
}

@article{zhang2024bootstrapping,
  title={Bootstrapping referring multi-object tracking},
  author={Zhang, Yani and Wu, Dongming and Han, Wencheng and Dong, Xingping},
  journal={arXiv preprint arXiv:2406.05039},
  year={2024}
}

@article{bai2025qwen2,
  title={Qwen2. 5-vl technical report},
  author={Bai, Shuai and Chen, Keqin and Liu, Xuejing and Wang, Jialin and Ge, Wenbin and Song, Sibo and Dang, Kai and Wang, Peng and Wang, Shijie and Tang, Jun and others},
  journal={arXiv preprint arXiv:2502.13923},
  year={2025}
}

@article{zheng2025group,
  title={Group sequence policy optimization},
  author={Zheng, Chujie and Liu, Shixuan and Li, Mingze and Chen, Xiong-Hui and Yu, Bowen and Gao, Chang and Dang, Kai and Liu, Yuqiong and Men, Rui and Yang, An and others},
  journal={arXiv preprint arXiv:2507.18071},
  year={2025}
}

@article{li2025cpany,
  title={CPAny: Couple With Any Encoder to Refer Multi-Object Tracking},
  author={Li, Weize and Du, Yunhao and Yin, Qixiang and Zhao, Zhicheng and Su, Fei and Liu, Daqi},
  journal={arXiv e-prints},
  pages={arXiv--2503},
  year={2025}
}

@inproceedings{long2019multi,
  title={Multi-adapter RGBT tracking},
  author={Long Li, Cheng and Lu, Andong and Hua Zheng, Ai and Tu, Zhengzheng and Tang, Jin},
  booktitle={Proceedings of the IEEE/CVF international conference on computer vision workshops},
  pages={0--0},
  year={2019}
}

@article{zhang2020object,
  title={Object tracking in RGB-T videos using modal-aware attention network and competitive learning},
  author={Zhang, Hui and Zhang, Lei and Zhuo, Li and Zhang, Jing},
  journal={Sensors},
  volume={20},
  number={2},
  pages={393},
  year={2020},
  publisher={MDPI}
}

@inproceedings{wang2020cross,
  title={Cross-modal pattern-propagation for RGB-T tracking},
  author={Wang, Chaoqun and Xu, Chunyan and Cui, Zhen and Zhou, Ling and Zhang, Tong and Zhang, Xiaoya and Yang, Jian},
  booktitle={Proceedings of the IEEE/CVF Conference on computer vision and pattern recognition},
  pages={7064--7073},
  year={2020}
}

@inproceedings{zhang2023efficient,
  title={Efficient rgb-t tracking via cross-modality distillation},
  author={Zhang, Tianlu and Guo, Hongyuan and Jiao, Qiang and Zhang, Qiang and Han, Jungong},
  booktitle={Proceedings of the IEEE/CVF conference on computer vision and pattern recognition},
  pages={5404--5413},
  year={2023}
}

@inproceedings{xiao2025cross,
  title={Cross-modulated Attention Transformer for RGBT Tracking},
  author={Xiao, Yun and Zhao, Jiacong and Lu, Andong and Li, Chenglong and Yin, Bing and Lin, Yin and Liu, Cong},
  booktitle={Proceedings of the AAAI Conference on Artificial Intelligence},
  volume={39},
  number={8},
  pages={8682--8690},
  year={2025}
}

@inproceedings{chen2025cross,
  title={Cross-view referring multi-object tracking},
  author={Chen, Sijia and Yu, En and Tao, Wenbing},
  booktitle={Proceedings of the AAAI Conference on Artificial Intelligence},
  volume={39},
  number={2},
  pages={2204--2211},
  year={2025}
}

@article{ma2025multi,
  title={Multi-Stage Cross-Modality Feature Interaction for RGB-Thermal Multi-Object Tracking},
  author={Ma, Jianbo and Luo, Hui and Niu, Shuaicheng and Zhao, Peilin and Liu, Yunfeng and Wei, Yuxing and Zhang, Jianlin},
  journal={IEEE Transactions on Circuits and Systems for Video Technology},
  year={2025},
  volume={},
  number={},
  pages={},
  doi={}
}

@article{ZHANG2025110984,
  title = {UniRTL: A universal RGBT and low-light benchmark for object tracking},
  journal = {Pattern Recognition},
  volume = {158},
  pages = {110984},
  year = {2025},
  issn = {0031-3203},
  author = {Lian Zhang and Lingxue Wang and Yuzhen Wu and Mingkun Chen and Dezhi Zheng and Liangcai Cao and Bangze Zeng and Yi Cai}
}

@article{hu2025deformle,
  title={Deformle Cross-Attention Trnsformer for Wekly Aligned RGB--T Pedestrin Detection},
  author={Hu, Yu and Chen, Xiaobo and Wang, Sheng and Liu, Luyang and Shi, Hengyang and Fan, Lihong and Tian, Jing and Liang, Jun},
  journal={IEEE Transactions on Multimedia},
  year={2025},
  publisher={IEEE}
}

@inproceedings{radford2021learning,
  title={Learning transferable visual models from natural language supervision},
  author={Radford, Alec and Kim, Jong Wook and Hallacy, Chris and Ramesh, Aditya and Goh, Gabriel and Agarwal, Sandhini and Sastry, Girish and Askell, Amanda and Mishkin, Pamela and Clark, Jack and others},
  booktitle={International conference on machine learning},
  pages={8748--8763},
  year={2021},
  organization={PmLR}
}

@article{liu2023visual,
  title={Visual instruction tuning},
  author={Liu, Haotian and Li, Chunyuan and Wu, Qingyang and Lee, Yong Jae},
  journal={Advances in neural information processing systems},
  volume={36},
  pages={34892--34916},
  year={2023}
}

@article{achiam2023gpt,
  title={Gpt-4 technical report},
  author={Achiam, Josh and Adler, Steven and Agarwal, Sandhini and Ahmad, Lama and Akkaya, Ilge and Aleman, Florencia Leoni and Almeida, Diogo and Altenschmidt, Janko and Altman, Sam and Anadkat, Shyamal and others},
  journal={arXiv preprint arXiv:2303.08774},
  year={2023}
}

@article{team2023gemini,
  title={Gemini: a family of highly capable multimodal models},
  author={Team, Gemini and Anil, Rohan and Borgeaud, Sebastian and Alayrac, Jean-Baptiste and Yu, Jiahui and Soricut, Radu and Schalkwyk, Johan and Dai, Andrew M and Hauth, Anja and Millican, Katie and others},
  journal={arXiv preprint arXiv:2312.11805},
  year={2023}
}

@inproceedings{chen2024internvl,
  title={Internvl: Scaling up vision foundation models and aligning for generic visual-linguistic tasks},
  author={Chen, Zhe and Wu, Jiannan and Wang, Wenhai and Su, Weijie and Chen, Guo and Xing, Sen and Zhong, Muyan and Zhang, Qinglong and Zhu, Xizhou and Lu, Lewei and others},
  booktitle={Proceedings of the IEEE/CVF conference on computer vision and pattern recognition},
  pages={24185--24198},
  year={2024}
}

@article{li2021lasher,
  title={LasHeR: A large-scale high-diversity benchmark for RGBT tracking},
  author={Li, Chenglong and Xue, Wanlin and Jia, Yaqing and Qu, Zhichen and Luo, Bin and Tang, Jin and Sun, Dengdi},
  journal={IEEE Transactions on Image Processing},
  volume={31},
  pages={392--404},
  year={2021},
  publisher={IEEE}
}

@inproceedings{zhang2022visible,
  title={Visible-thermal UAV tracking: A large-scale benchmark and new baseline},
  author={Zhang, Pengyu and Zhao, Jie and Wang, Dong and Lu, Huchuan and Ruan, Xiang},
  booktitle={Proceedings of the IEEE/CVF conference on computer vision and pattern recognition},
  pages={8886--8895},
  year={2022}
}

@article{zhu2025visible,
  title={Visible--thermal multiple object tracking: Large-scale video dataset and progressive fusion approach},
  author={Zhu, Yabin and Wang, Qianwu and Li, Chenglong and Tang, Jin and Gu, Chengjie and Huang, Zhixiang},
  journal={Pattern Recognition},
  volume={161},
  pages={111330},
  year={2025},
  publisher={Elsevier}
}

@article{ge2021yolox,
  title={Yolox: Exceeding yolo series in 2021},
  author={Ge, Zheng and Liu, Songtao and Wang, Feng and Li, Zeming and Sun, Jian},
  journal={arXiv preprint arXiv:2107.08430},
  year={2021}
}

@inproceedings{zhang2022bytetrack,
  title={Bytetrack: Multi-object tracking by associating every detection box},
  author={Zhang, Yifu and Sun, Peize and Jiang, Yi and Yu, Dongdong and Weng, Fucheng and Yuan, Zehuan and Luo, Ping and Liu, Wenyu and Wang, Xinggang},
  booktitle={European conference on computer vision},
  pages={1--21},
  year={2022},
  organization={Springer}
}

@article{kalman1960new,
  title={A new approach to linear filtering and prediction problems},
  author={Kalman, Rudolph Emil},
  year={1960}
}

@article{kuhn1955hungarian,
  title={The Hungarian method for the assignment problem},
  author={Kuhn, Harold W},
  journal={Naval research logistics quarterly},
  volume={2},
  number={1-2},
  pages={83--97},
  year={1955},
  publisher={Wiley Online Library}
}

@article{shao2024deepseekmath,
  title={Deepseekmath: Pushing the limits of mathematical reasoning in open language models},
  author={Shao, Zhihong and Wang, Peiyi and Zhu, Qihao and Xu, Runxin and Song, Junxiao and Bi, Xiao and Zhang, Haowei and Zhang, Mingchuan and Li, YK and Wu, Yang and others},
  journal={arXiv preprint arXiv:2402.03300},
  year={2024}
}

@misc{liu2024llavanext,
  title={Llavanext: Improved reasoning, ocr, and world knowledge},
  author={Liu, Haotian and Li, Chunyuan and Li, Yuheng and Li, Bo and Zhang, Yuanhan and Shen, Sheng and Lee, Yong Jae},
  year={2024}
}

\section{Appendix}
\label{sec:Appendix}
\quad In this supplementary material, we provide additional more details about our research. Specifically, \cref{sec:Comparison with Existing Datasets} offers a further comparison between the RefRT dataset and other related datasets. \cref{sec:Analysis of Rewards} presents a comprehensive analysis of the proposed reward design. \cref{sec:Prompt Designs for MLLMs} presents the prompt designs for the MLLMs used by the RTrack framework. \cref{sec:Prompt robustness analysis.} presents a robustness analysis of RTrack under different prompts. \cref{sec:Implementation Details} summarizes the detailed experimental configurations and parameter settings.  \cref{sec:Visualization results} presents additional visualization results. \cref{sec:Ground-Truth Results} presents more examples of ground-truth annotations. 

\subsection{Comparison with Existing Datasets}
\label{sec:Comparison with Existing Datasets}
\quad We compare the proposed RefRT dataset with several representative datasets in the fields of RGBT tracking and RMOT, as summarized in \cref{tab:dataset_compare}. The comparison covers several key dimensions, including scene diversity, modality configuration, annotation richness, availability of language descriptions and task relevance. As shown in \cref{tab:dataset_compare}, existing datasets either provide RGBT image pairs without language descriptions or offer referring multi-object tracking annotations without thermal modality. Consequently, no existing dataset supports RMOT under the RGBT setting, which limits the development and evaluation of RT-RMOT methods. The RefRT dataset fills this gap as the first multi-object tracking dataset that simultaneously provides RGBT modalities and high-quality language descriptions. It covers a diverse set of mixed urban and road scenes and includes 14 object categories. The dataset offers 388 language descriptions involving 1,354 tracking targets, with each description capturing target behaviors, scene context, and inter-object interactions. Moreover, the precisely aligned RGBT image frames further support robust tracking in scenarios with low light, low visibility, and low visual contrast. Compared with existing RGB modality datasets or those lacking language descriptions, the RefRT dataset offers significant advantages in environmental robustness, task generalization, and semantic awareness. It provides a high quality data foundation for advancing RGB-Thermal Referring Multi-Object Tracking research in complex environments.

\begin{table*}[tb]
\setlength{\abovecaptionskip}{0.6mm}
\setlength{\belowcaptionskip}{0mm}
\centering
\caption{\textbf{Comparison with Existing Datasets.} The \textbf{Desc.} indicates the number of language descriptions, and \textbf{Instance/Desc.} represents the average number of target instances per description. Check marks indicate whether a dataset supports Multi-Object Tracking, includes RGBT modalities data, and contains real-world scenarios. The comparison results indicate that the RefRT dataset fulfills the combined requirements of RGBT modalities inputs, diverse object categories, and comprehensive semantic annotations.}
\label{tab:dataset_compare}
\renewcommand{\arraystretch}{1.3}

\resizebox{\linewidth}{!}{
\begin{tabular}{*{11}{c}}
\hline
\rowcolor{gray!30}
Dataset & Source & Videos & Scene & Desc. & Instance & Instance/Desc. & Bbox & Multi-Object & RGBT & Real-world \\
\hline
RGBT234 & IJCV2019 & 234 & Urban, campus & 234 & – & – & 234 & × & \checkmark & \checkmark \\
GTOT & CVPR2016 & 50 & Urban, campus & – & – & – & 15.8k & × & \checkmark & \checkmark  \\
RGBT-Tiny & arXiv 2024 & 115 & Port, urban roads & – & – & – & 1.2M & × & \checkmark & \checkmark  \\
Refer-KITTI & CVPR2023 & 18 & Urban, road mixed & 818 & 8.7k & 10.7 & 0.36M & \checkmark & × & \checkmark  \\
Refer-KITTI-V2 & arXiv 2024 & 21 & Urban, highways, etc. & 9.8k & 65.4k & 607 & 3.06M & \checkmark & × & \checkmark  \\
Refer-Dance & CVPR2024 & 65 & Mainly indoor & 1.9k & 650 & 0.34 & 0.55M & \checkmark & × & \checkmark  \\
Refer-UE-City & arXiv 2024 & 14 & Urban, streets, etc. & 714 & 7.3k & 10.3 & 0.56M & \checkmark & × & \checkmark  \\
CRTrack & AAAI2025 & 41 & Urban, outdoor & 221 & 344 & 1.56 & 0.79M & \checkmark & × & \checkmark  \\
\textbf{RefRT} & \textbf{Ours} & 72 & Urban, road mixed scenes & 388 & 1354 & 3.48 & 0.27M & \checkmark & \checkmark & \checkmark \\
\hline
\end{tabular}
}
\end{table*}

\subsection{Analysis of Rewards}
\label{sec:Analysis of Rewards}
\quad For the two reward functions proposed in our method, namely the Structured Output Reward and the Comprehensive Detection Reward, we provide a detailed analysis of their functional roles and practical effectiveness.

\textbf{(1) Structured Output Reward.} This reward not only enforces a standardized output format, ensuring that the target bounding box information can be reliably parsed from the generated results, but also constrains the length of the output. An excessively long response often indicates that the model has produced redundant reasoning or unnecessary explanations, drifting away from the core detection task; conversely, an overly short response may lead to missing key information, preventing the model from fully conveying its localization results or reasoning process. Therefore, by jointly constraining both format and length, this reward guides the model to generate structured and sufficiently informative detection outputs, thereby improving stability and effectiveness during the reinforcement learning process.

\textbf{(2) Comprehensive Detection Reward.} This reward adopts a staged design in which the Output Encouragement Reward is used first to enhance the model’s exploration capability, and the Precision Detection Reward is subsequently applied to strengthen its detection precision.

The Output Encouragement Reward adopts a coverage-first principle, aiming to guide the model to discover and cover as many ground-truth targets as possible in the early training stage. After applying Hungarian matching, a higher number of matched ground-truth instances indicates that the model is exploring a broader and more diverse area and detecting more targets, which is essential for multi-object tracking. To promote this behavior, the reward provides positive feedback for increasing ground-truth coverage, preventing the model from collapsing into a conservative strategy that outputs only a few high-confidence boxes. This coverage-oriented design helps reduce severe missed detections during cold start and establishes a solid foundation for improving detection precision in later training. It expands the model’s exploration space and leads to a more complete understanding of target distributions, thereby benefiting the subsequent precise-detection stage.
\begin{equation}
R_{OER}(B_{det}, B_{gt}) = \alpha \cdot MatchedGT + \beta \cdot IoU_{score}
\label{eq:oer}
\end{equation}
where \textit{MatchedGT} denotes the number of valid ground-truth boxes that are successfully matched with predictions.

The Precision Detection Reward is designed to enhance the model’s localization accuracy once its detection coverage has been established. Its core idea is to jointly constrain prediction quality and output quantity, guiding the model to produce more accurate and more compact detection results. The first term of the reward is based on the IoU between matched predictions and ground-truth targets, normalized by the number of predicted boxes. This prevents the model from obtaining higher rewards simply by generating an excessive number of high-confidence candidate boxes. As a result, the model must rely on genuinely precise spatial localization rather than stacking redundant boxes around the ground-truth region and depending on Hungarian matching to filter them afterward. To prevent the model from focusing solely on high-precision detections while neglecting other targets, the second term measures the proportion of covered ground truths, encouraging the model to detect all GT instances as completely as possible. By combining these two terms, the reward achieves a balance between precision and coverage, allowing the model to maintain high recall while gradually reducing redundant predictions and significantly improving localization accuracy. Through this mechanism, the model gradually shifts in the later stages of training from simply ensuring that all targets are covered to consistently achieving precise and reliable localization for every target. This progression leads to more stable behavior and significantly improves the overall quality of the detection results.
\begin{equation}
R_{PDR}(B_{\text{det}}, B_{\text{gt}}) =
\frac{IoU_{score}}{(N_{\text{det}})^{\gamma}} +
\frac{\lambda \cdot MatchedGT}{N_{\text{gt}}}
\end{equation}
where \( IoU_{\text{score}} \) and \textit{MatchedGT} have the same meanings as defined in Eq.~(\ref{eq:oer}). Here, $N_{\text{det}}$ and $N_{\text{gt}}$ denote the numbers of predicted and ground-truth boxes.

\subsection{Prompt Designs for MLLMs}
\label{sec:Prompt Designs for MLLMs}
\quad To enable the multimodal large language model (MLLM) in the RTrack framework to produce the desired bounding box outputs, it is necessary to design tailored prompts for each model. Our prompt designs for the MLLMs used in the RTrack framework are presented below:

\textbf{(1) Prompt Design for Qwen2.5-VL:} We input the prompt as "You are a Visual Language Model specifically designed for paired and perfectly aligned RGB + thermal images. Please utilize the information from both modes simultaneously and detect all targets that match: " + language description + "in the image and output their coordinates with \texttt{[x1,y1,x2,y2]} format. First output the thinking process in \texttt{<think></think>} tags and then output the final answer in \texttt{<answer></answer>} tags. Note that the \texttt{<answer></answer>} tags should not contain any text, only the coordinates in the \texttt{[x1,y1,x2,y2]} format."

\textbf{(2) Prompt Design for InternVL:} We input the prompt as "The input consists of an aligned pair of RGB and thermal (T) images. Please combine information from both modalities, follow the user’s text instruction, and identify all target objects that match. Please output the final bounding box coordinates in the format of \texttt{[x1,y1,x2,y2]}, and present them in the form of a list. User instruction is:"  + language description.

\textbf{(3) Prompt Design for LLaVA and LLaVA-NEXT:} We input the prompt as "You are a VLM (Visual Language Model) specifically designed for paired and perfectly aligned RGB + thermal images. Please utilize the information from both modes simultaneously and detect all " + language description + " in the image and output their coordinates with \texttt{[x1,y1,x2,y2]} format."

\subsection{Prompt robustness analysis.} 
\label{sec:Prompt robustness analysis.}
\quad We validate its robustness from three aspects: prompt complexity, description position, and output coordinate format. All experiments use the same RTrack model after fine-tuning. As shown in \cref{tab:prompt_ablation}, although these variations lead to slight performance fluctuations, RTrack still maintains good performance.

\begin{table}[t]
\setlength{\abovecaptionskip}{-0.2mm}
\centering
\caption{Ablation study on different prompt designs in RTrack.}
\label{tab:prompt_ablation}
\resizebox{1.0\linewidth}{!}{
    \begin{tabular}{lccccccccc}
    \toprule
    \textbf{Prompt Design} 
    & HOTA↑ & DetA↑ & AssA↑ & DetRe↑ & DetPr↑ & AssRe↑ & AssPr↑ & LocA↑ \\
    \midrule
    Short Prompt & 13.92 & 11.99 & 17.42 & 17.92 & 24.75 & 18.30 & \textbf{82.87} & 75.64 \\
    Description-First Prompt  & 13.00 & 11.10 & 16.25 & \textbf{20.21} & 18.75 & 17.05 & 81.68 & \textbf{75.65} \\
    No Coordinate Example Prompt & \textbf{16.29} & 12.26 & \textbf{22.90} & 17.95 & \textbf{25.85} & \textbf{24.49} & 81.25 & 75.35 \\
    \rowcolor{gray!30}
    \textbf{Base Prompt (Ours)} & 15.53 & \textbf{12.39} & 20.79 & 20.15 & 22.78 & 22.02 & 81.99 & 75.53 \\
    \bottomrule
    \end{tabular}
}
\end{table}

\begin{table}[tb]
    \setlength{\abovecaptionskip}{0.6mm}
    \caption{Hyperparameter configuration used in the RL fine-tuning process.}
    \label{tab:hyperparameters}
    \renewcommand{\arraystretch}{1.3}
    \centering
    \resizebox{\linewidth}{!}{
    \begin{tabular}{ccccccccccccccccccc}
    \hline
    \rowcolor{gray!30}
    Batch Size & Epochs & Learning Rate & LoRA-$r$ & LoRA-$\alpha$ & LoRA Dropout & Generation Num & Max Len & Beta & Epsilon & $Scale_{\max}$ & $L_{\min}$ & $L_{\max}$ & $L_{low}$ & $L_{high}$ & $\alpha$ & $\beta$ & $\gamma$ & $\lambda$ \\
    \hline
    1 & 1 & $1\times10^{-5}$ & 8 & 16 & 0.05 & 4 & 512 & 0.001 & $1\times10^{-3}$ & 3.0 & 80 & 600 & 140 & 200 & 0.5 & 1 & 0.5 & 2 \\
    \hline
    \end{tabular}
    }
\end{table}

\subsection{Implementation Details}
\label{sec:Implementation Details}
\quad To ensure the reproducibility of our experimental results, we summarize all hyperparameter configurations used throughout our study in Table~\ref{tab:hyperparameters}. These include optimization settings, fine-tuning parameters, generation configurations and the detailed hyperparameters associated with the reward mechanism. Furthermore, for the Comprehensive Detection Reward, we apply the Output Encouragement Reward in the first phase of training to enhance the model’s exploration capability, and then switch to the Precision Detection Reward in the later phase to strengthen the model’s ability to utilize information accurately.

\subsection{Visualization Results}
\label{sec:Visualization results}
\begin{figure}[tb]
    \setlength{\abovecaptionskip}{0.8mm}   
    \setlength{\belowcaptionskip}{-2.4mm}  
    \centering
    \includegraphics[width=\textwidth]{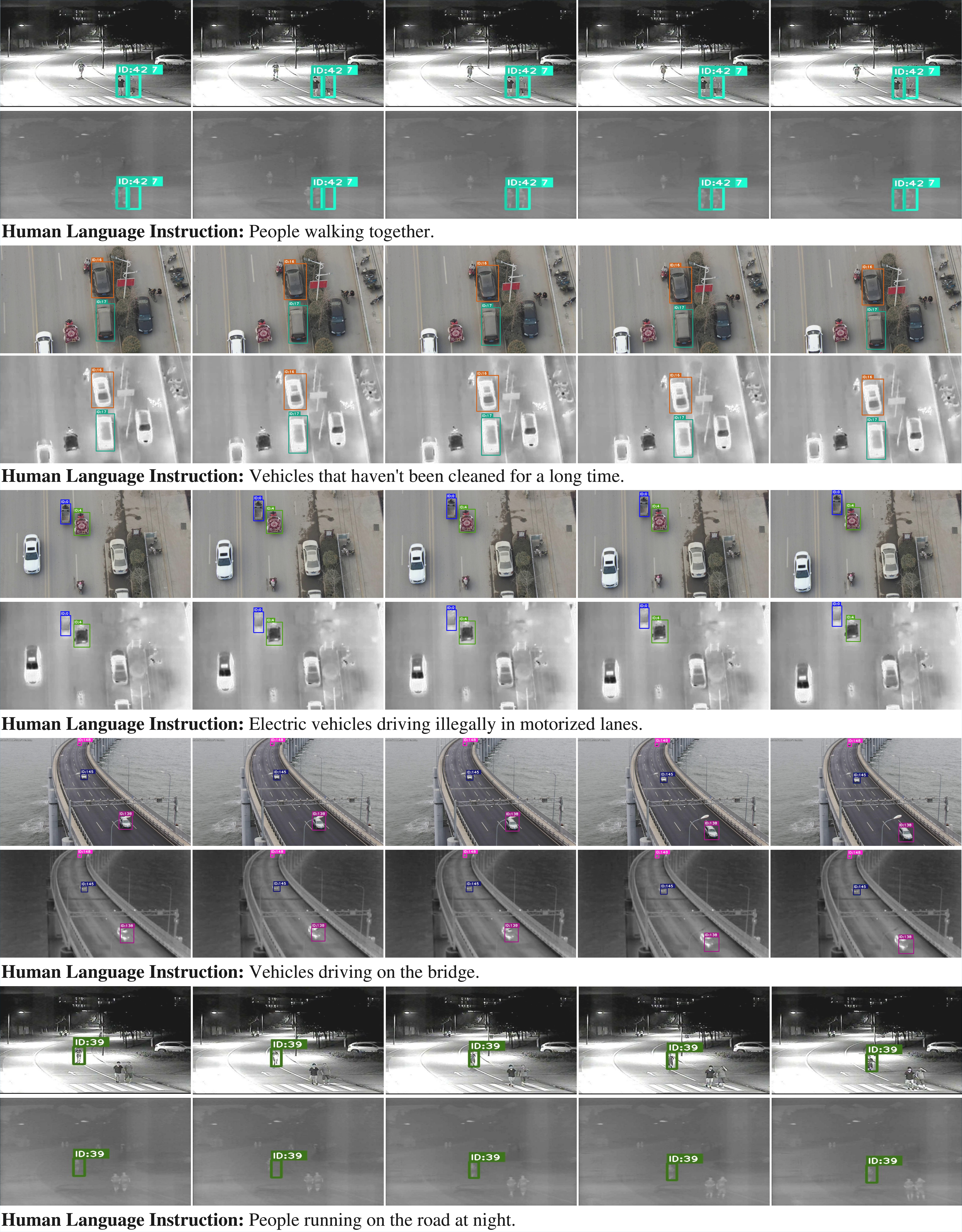}
    \caption{\textbf{More qualitative results} of our RTrack framework on the test set of the RefRT dataset . It is worth noting that our framework can perform long-term tracking, rather than tracking only in the initial frame.}
    \label{fig:Visualization_supplementary}
\end{figure}

\quad We visualize additional results of the RTrack framework on the RefRT dataset test set. As shown in \cref{fig:Visualization_supplementary}, the figure provides a broader set of qualitative examples that illustrate the performance of the framework. These results indicate that RTrack maintains strong tracking capability even in complex and dynamic environments. For example, in the first example of \cref{fig:Visualization_supplementary}, given the instruction “People walking together,” the model correctly identifies and tracks the two adjacent, slow-moving individuals in the scene rather than the faster-moving one. This demonstrates the framework’s ability to interpret subtle semantic cues in the language description and to distinguish targets that may appear visually similar but differ in behavior. In the fourth example, with the description “Vehicles driving on the bridge,” the last vehicle in the sequence appears small and blends into the road surface in the RGB image. Nonetheless, the model successfully detects and tracks it. This case shows that RTrack can rely on RGB cues to understand scene-level semantics while using IR imagery to capture target contours, thereby achieving complementary cross-modal tracking. These qualitative results demonstrate that RTrack effectively integrates RGBT information to infer the targets specified by natural-language instructions and maintain stable long-term tracking. The findings confirm that RTrack is a reliable and robust framework capable of supporting real-world RT-RMOT task.

\subsection{Ground-Truth Results}
\label{sec:Ground-Truth Results}
\begin{figure}[tb]
    \setlength{\abovecaptionskip}{0.8mm}  
    \setlength{\belowcaptionskip}{-2.4mm} 
    \centering
    \captionsetup{singlelinecheck=false}
    \captionsetup{justification=raggedright}
    \includegraphics[width=\textwidth]{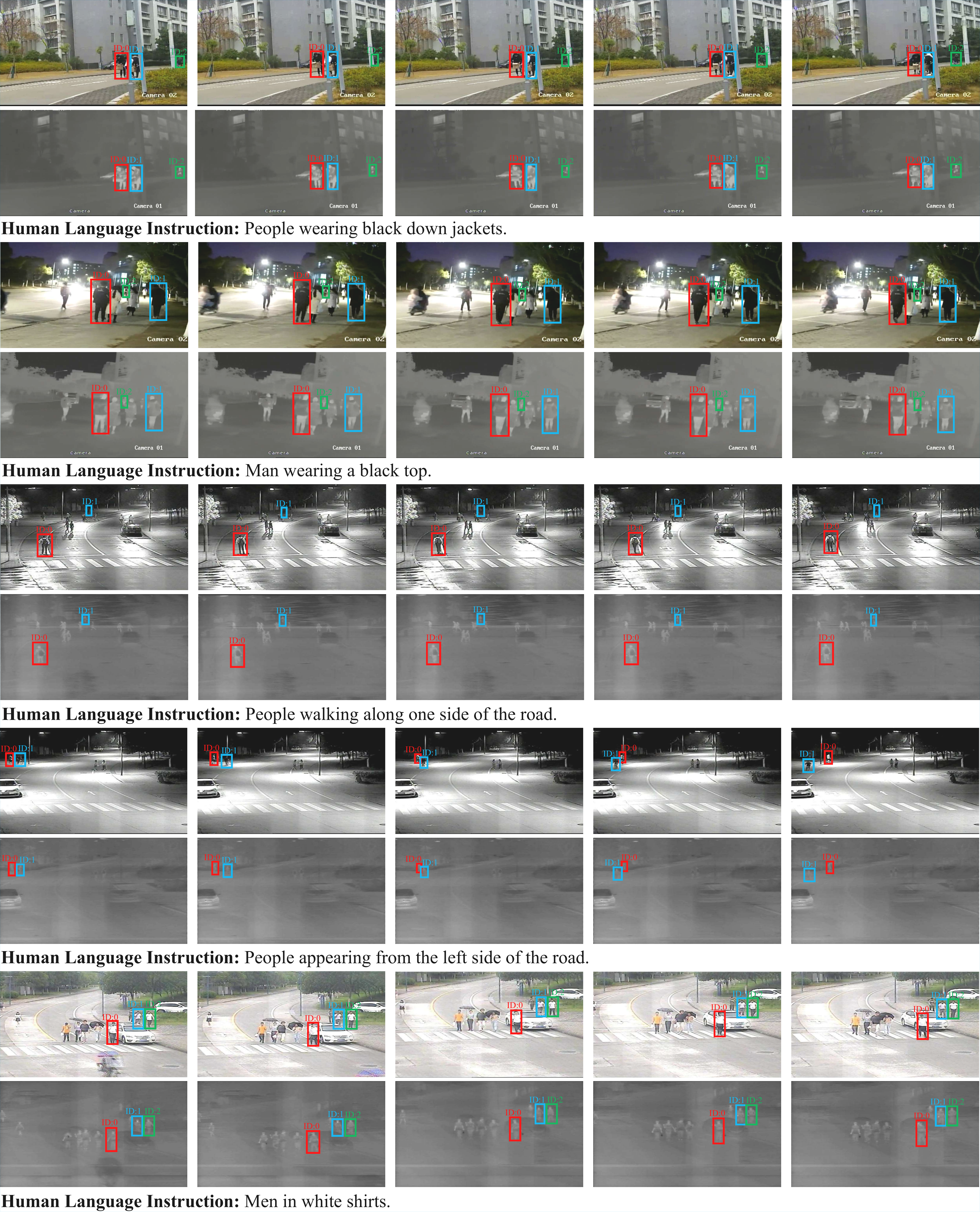}
    \caption{\textbf{Ground-Truth Results} in the RefRT dataset provide diverse, RGBT aligned real-world data.}
    \label{fig:gt_supplementary}
\end{figure}

\quad In constructing the RefRT dataset, we leveraged the complementary sensing properties of RGB and IR imagery to produce precisely aligned RGBT annotations. This carefully designed RGBT modality setting provides a more comprehensive and reliable representation of the scene than relying on either modality alone. It enables the system to localize referred targets with greater robustness under complex environmental conditions as well as diverse and potentially ambiguous language descriptions. As presented in \cref{fig:gt_supplementary}, the contribution of the infrared modality becomes evident across several representative cases. In the first example, the target described as “People wearing black down jackets”, ID 2, nearly disappears within the dark RGB background, making it difficult to identify through visible cues alone. In the second example, “Men wearing a black top”, target ID 2 is partially occluded and surrounded by other individuals, increasing the chance that it may be overlooked in RGB frames. Infrared imagery offers clear and stable contours that support reliable localization. After the target position is estimated, RGB appearance information further assists in matching the correct instance to the language description. Similarly, in the third case shown in \cref{fig:gt_supplementary}, “People walking along one side of the road,” the targets appear visually indistinct in the RGB images, complicating candidate selection. In contrast, the infrared modality reveals the thermal differences between the targets and the background, which collectively offer distinctive cues that effectively facilitate target discrimination and tracking.

\quad RGB images capture rich appearance cues and high-level semantic information, whereas thermal images provide essential support for reliable target differentiation and stable tracking under complex conditions. By leveraging aligned RGBT image pairs, RefRT delivers a diverse, highly consistent, and semantically coherent representation of the RGBT modality. The integration of RGBT information establishes a versatile foundation for advancing research on the RT-RMOT task and for developing models capable of robust multimodal perception.

\end{document}